\documentclass[sigconf]{acmart}

\renewcommand\footnotetextcopyrightpermission[1]{} % removes footnote with conference information in first column
\usepackage{multirow}
\usepackage{booktabs}
\usepackage{lipsum}
\usepackage{booktabs}
\usepackage{multicol}
\usepackage{multirow}
\usepackage{color, colortbl}
\usepackage{nicematrix}
\usepackage{xcolor}
\usepackage{adjustbox}

\newcommand\egno{\textit{e.g.}}
\newcommand\ieno{\textit{i.e.}}

\newcommand{\etal}{{\textit{et al}}.\@ }
\definecolor{Gray}{gray}{0.9}
\newcommand{\CC}{\cellcolor{gray!20}}

\AtBeginDocument{% 
  \providecommand\BibTeX{{%
    \normalfont B\kern-0.5em{\scshape i\kern-0.25em b}\kern-0.8em\TeX}}}

\begin{document}

\title{Can SAM Boost Video Super-Resolution?}
% \title{Light Field Non-Local Spatial Super-Resolution Using Dcoupling Matching}
% \author{Anonymous  submission}
% \email{xxx}

% \author{Anonymous submission}
% \affiliation{%

\author{Zhihe Lu*$^1$, Zeyu Xiao*$^{1,2}$, Jiawang Bai*$^{1,3}$, Zhiwei Xiong$^2$, and Xinchao Wang$^1$}
 \affiliation{
  \institution{$^1$National University of Singapore, $^2$University of Science and Technology of China, $^3$Tsinghua University}
  \authornote{equal contribution.}
  \country{}
 }
% \renewcommand{\shortauthors}{Anonymous Author,~\textit{et al.}}

% \renewcommand{\shortauthors}{Trovato and Tobin,~\textit{et al.}}

\begin{abstract}
The primary challenge in video super-resolution (VSR) is to handle large motions in the input frames, which makes it difficult to accurately aggregate information from multiple frames.
Existing works either adopt deformable convolutions or estimate optical flow as a prior to establish correspondences between frames for the effective alignment and fusion.
However, they fail to take into account the valuable semantic information that can greatly enhance it; and flow-based methods heavily rely on the accuracy of a flow estimate model, which may not provide precise flows given two low-resolution frames.

In this paper, we investigate a more robust and semantic-aware prior for enhanced VSR by utilizing the Segment Anything Model (SAM), a powerful foundational model that is less susceptible to image degradation.
To use the SAM-based prior, we propose a simple yet effective module -- {\bf S}AM-guid{\bf E}d refin{\bf E}ment {\bf M}odule (SEEM), which can enhance both alignment and fusion procedures by the utilization of semantic information.
This light-weight plug-in module is specifically designed to {not only} leverage the attention mechanism for the generation of semantic-aware feature {but also} be easily and seamlessly integrated into existing methods.
Concretely, we apply our SEEM to two representative methods, EDVR and BasicVSR, resulting in consistently improved performance with minimal implementation effort, on three widely used VSR datasets: Vimeo-90K, REDS and Vid4.
More importantly, we found that the proposed SEEM can advance the existing methods in an efficient tuning manner, providing increased flexibility in adjusting the balance between performance and the number of training parameters.
Code will be open-source soon.
\end{abstract}


% \ccsdesc[500]{Computing methodologies~Computational photography}
% \ccsdesc{Reconstruction}



\maketitle

\section{Introduction}
Video super-resolution (VSR) \cite{fuoli2019efficient,haris2019recurrent,huang2015bidirectional,huang2017video,isobe2020video} is a fundamental low-level vision task that aims to create a more detailed and visually appealing video from a low-resolution version, which has gained increasing popularity in the multimedia field thanks to its practical applications, \egno, video surveillance \cite{zhang2010super}, high-definition television \cite{goto2014super}, and satellite imagery \cite{deudon2020highres,luo2017video}.
The main challenge of VSR is information aggregation given multiple frames with large motions and occluded regions.
In general, the aggregation process consists of two key steps: alignment and fusion, for which two types of main-stream methods have been proposed: sliding window-based and bidirectional recurrent structure-based.
Concretely, one common practice used in sliding window-based methods is deformable convolutions (DCN) \cite{tian2020tdan,wang2019edvr}, which aims to align multiple features implicitly.
In contrast, optical flow estimate \cite{caballero2017real,xue2019video} is adopted in bidirectional recurrent structure-based methods to explicitly build the correspondences between frames to enhance alignment and fusion.

\begin{figure}
    \centering
    \includegraphics[width=0.40\textwidth]{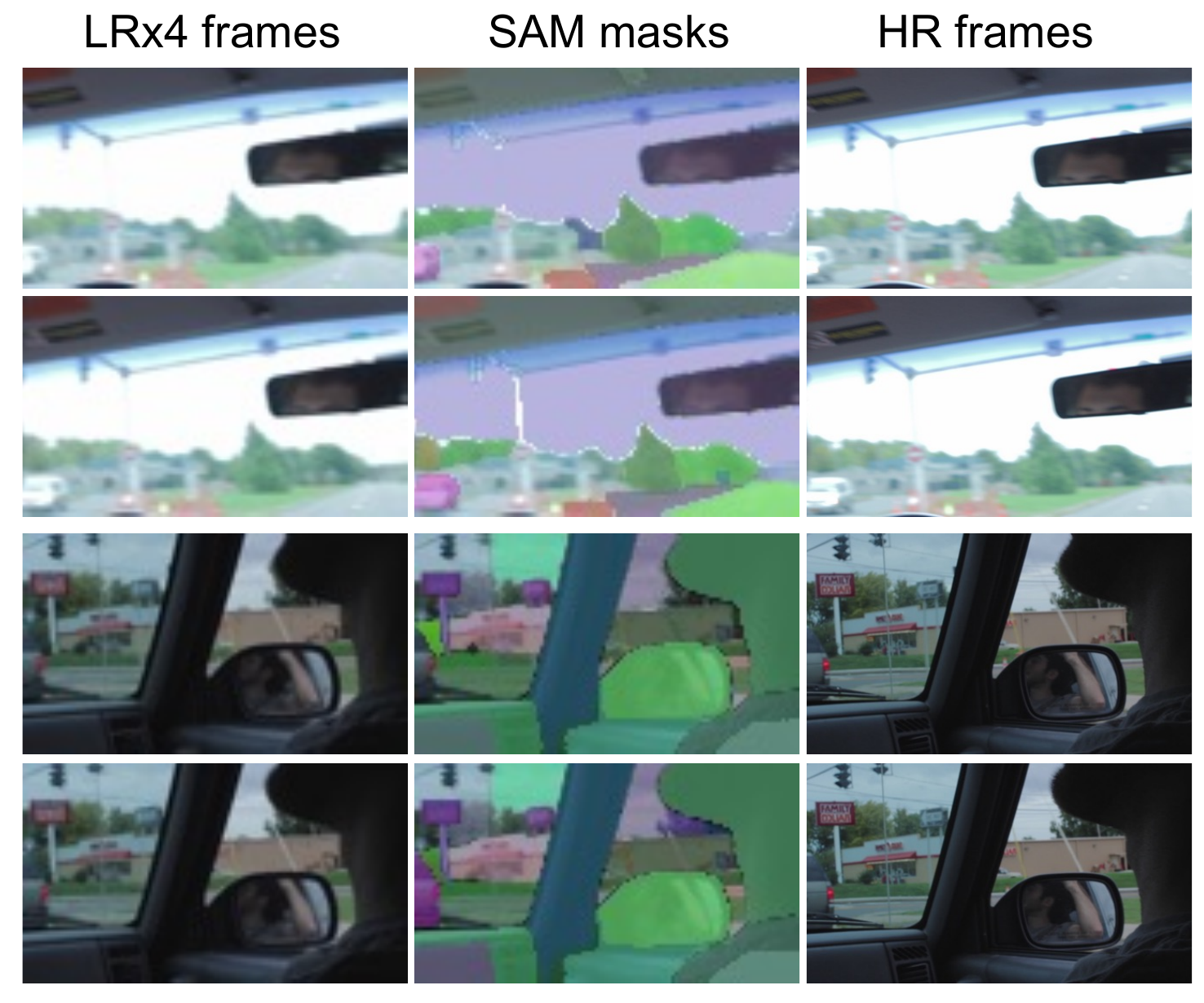}
    %\vspace{-4mm}
    \caption{Illustration of SAM's robustness on low-resolution video frames. It shows that SAM can accurately segment objects with fine edges, even in low-resolution images. This grounds our motivation to use the SAM-based prior for better video super-resolution.}
    \label{fig:lr_sam}
    %\vspace{-4mm}
\end{figure}

Despite the promising aggregation performance achieved by the two mentioned techniques, they encounter evident limitations.
Specifically, DCN-based methods \cite{tian2020tdan} often implicitly align the features of different frames by a normal regression supervision.
No extra supplementary information as guidance makes this alignment extremely challenging, often leading to a sub-optimal solution.
In contrast, the estimated optical flow \cite{caballero2017real,xue2019video} as the prior knowledge can improve the alignment as well as the fusion by guiding spatial warping.
However, the flow-based methods inevitably face two drawbacks: (i) they heavily rely on the performance of a flow estimate model, which may produce low-quality flows when dealing with degraded, \egno, low-resolution, images \cite{wang2019edvr}; (ii) the optimal flow is a means of modeling motions between frames, but lacks the semantic information, which is naturally helpful for correspondence establishment between the objects existing in continual frames.
This encourages us to explore more robust and semantic-aware knowledge to facilitate existing methods.

In this paper, we for the first time investigate the effectiveness of a semantic-aware prior extracted from a powerful pre-trained foundation model -- segment anything model (SAM) \cite{kirillov2023segment} for enhanced VSR.
Using this SAM-based prior is based on the observation that SAM is robust to the image degradation as shown in Figure \ref{fig:lr_sam} due to its impressive representation learned on 1 billion masks and 11 million images with 636 million parameters (ViT-H \cite{dosovitskiy2020ViT}). 
Specifically, we obtain this SAM-based prior by simply feeding a degraded image, \ieno, a low-resolution image in our case, to SAM.
SAM can then yield the masks for all possible objects included in the image.
However, given these masks, it still needs a specific design to utilize them effectively.

\begin{figure*}[ht]
    \centering
    \includegraphics[width=0.75\textwidth]{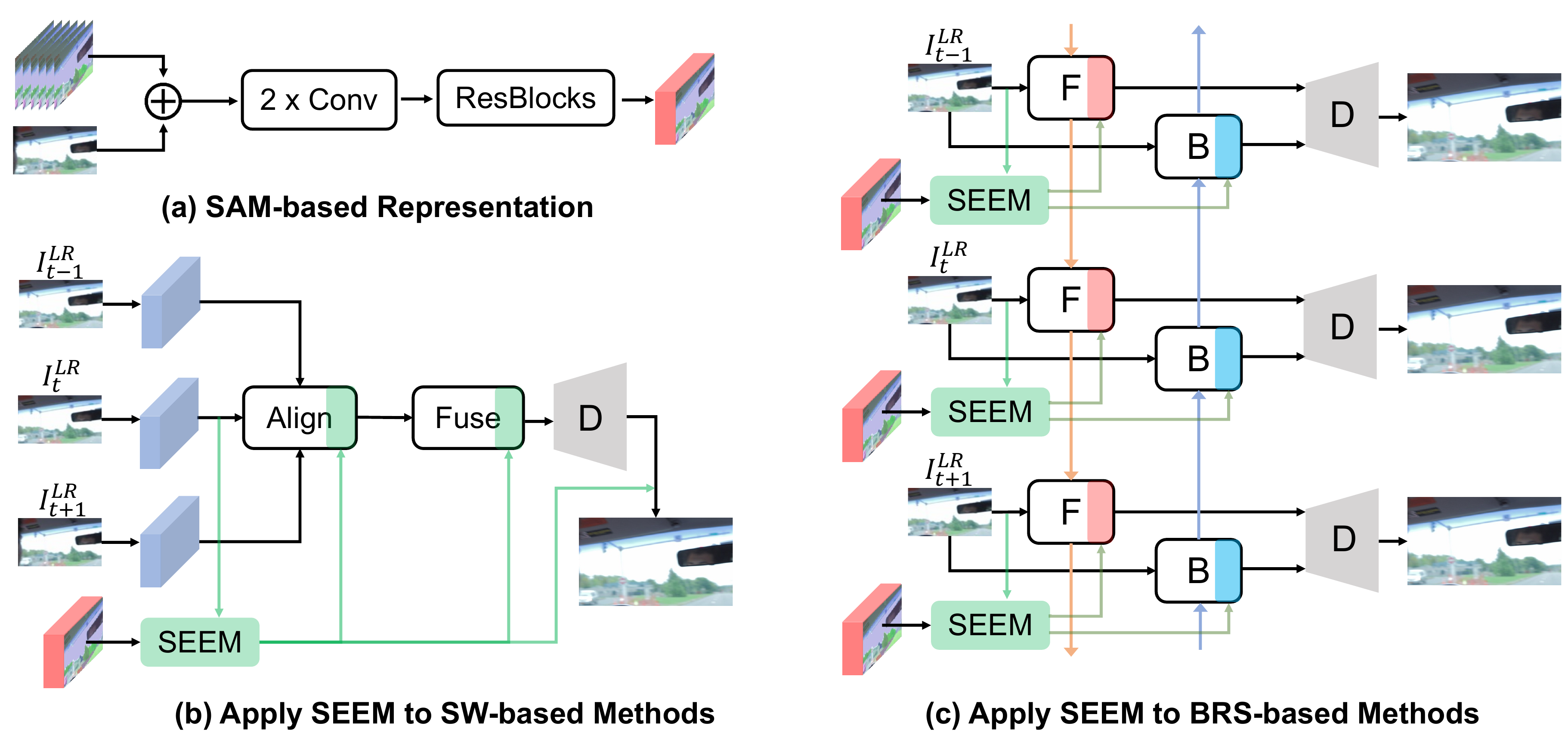}
    %\vspace{-4mm}
    \caption{Overview of the framework. (a) We illustrate how to obtain SAM-based representation. (b) We apply the proposed SAM-guided refinement module (SEEM) to the sliding-window based method. (c) We apply SEEM to the bidirectional recurrent structure based method. ``$\mathbf{F}$'' and ``$\mathbf{B}$'' are the forward- and backward- propagation. $\oplus$ represents the concatenation operation. The detailed structure of SEEM is shown in Figure \ref{fig:seem}.}
    \label{fig:framework}
    %\vspace{-4mm}
\end{figure*}

To leverage the SAM-based prior, we propose a novel plug-in module -- {\bf S}AM-guid{\bf E}d refin{\bf E}ment {\bf M}odule (SEEM), which can enhance both alignment and fusion procedures by the utilization of semantic information.
To be specific, the SEEM is designed to combine the SAM-based representation and the feature of the current frame to generate the semantic-aware feature.
This is achieved by leveraging the attention mechanism and several feature mapping operations.
Regarding the typical architectures of existing methods, the obtained semantic-aware feature can then be used in different ways for performance enhancement.
For sliding window-based methods, we introduce our SEEM to improve three procedures, \ieno, alignment, fusion and reconstruction, as shown in Figure \ref{fig:framework} (b).
In contrast, our SEEM is introduced to bidirectional branches for better feature warping and refinement in recurrent structure-based methods (Figure \ref{fig:framework} (c)).
It is worth noting that our SEEM exhibits high flexibility in being adopted by various methods thanks to its model-agnostic design, allowing it to be easily integrated into these methods without requiring any modifications to their original architectures.
More importantly, the proposed SEEM can advance the existing methods in both full fine-tuning and efficient tuning manners.

We summarize our contributions as follows.
% {\bf (i)} In light of the limitations of existing video super-resolution (VSR) works, we conduct a study to explore the impact of a more robust and semantic-aware prior. 
% Notably, we are the first to incorporate prior knowledge extracted from the segment anything model (SAM), the largest foundation model for segmentation to date, to improve the quality of VSR.
% {\bf (ii)} We propose a novel {\bf S}AM-guid{\bf E}d refin{\bf E}ment {\bf M}odule, referred to as SEEM, which serves as a lightweight plug-in that can be seamlessly integrated into existing methods to enhance their performance.
% {\bf (iii)} We introduce SEEM to sliding window-based method -- EDVR for enhancing various procedures; and to bidirectional recurrent structure-based methods -- BasicVSR for semantic-aware bidirectional interaction.
% {\bf (iv)} We conduct extensive experiments on three widely used VSR datasets: Vimeo-90K, REDS and Vid4, demonstrating the superior performance against the baseline methods.
% Further analysis also shows that our SEEM can advance the existing methods in a parameter-efficient manner, \ieno, only the parameters of SEEM are trainable during fine-tuning.
\begin{itemize}
\item In light of the limitations of existing video super-resolution (VSR) works, we conduct a study to explore the impact of a more robust and semantic-aware prior. 
Notably, we are the first to incorporate prior knowledge extracted from the segment anything model (SAM), the largest foundation model for segmentation to date, to improve the quality of VSR.
\item We propose a novel {\bf S}AM-guid{\bf E}d refin{\bf E}ment {\bf M}odule, referred to as SEEM, which serves as a lightweight plug-in that can be seamlessly integrated into existing methods to enhance their performance.
\item We introduce SEEM to sliding window-based method -- EDVR for enhancing various procedures; and to bidirectional recurrent structure-based methods -- BasicVSR for semantic-aware bidirectional interaction.
\item We conduct extensive experiments on three widely used VSR datasets: Vimeo-90K, REDS and Vid4, demonstrating the superior performance against the baseline methods.
Further analysis also shows that our SEEM can advance the existing methods in a parameter-efficient manner, \ieno, only the parameters of SEEM are trainable during fine-tuning.
\end{itemize}  

\section{Related Work}
\paragraph{Video Super-resolution}
Existing video super-resolution (VSR) methods aim to restore high-resolution (HR) frames by extracting more temporal information using sliding-window and recurrent structures.
Specifically, the methods using sliding-window structures, such as 3DSRNet \cite{kim20183dsrnet}, TDAN \cite{tian2020tdan}, and EDVR \cite{wang2019edvr}, typically recover HR frames from adjacent low-resolution (LR) frames by the dynamical offsets prediction of sampling convolution kernels within a sliding window.
To align temporal features, some specifically designed architectures or techniques, \egno, 3D convolution~\cite{kim20183dsrnet}, optical flow \cite{kim2018spatio, tao2017detail}, and deformable convolution \cite{tian2020tdan,wang2019edvr}, are adopted.
For example, 3DSRNet \cite{kim20183dsrnet} employs 3D convolution to extract temporal features, STTN \cite{kim2018spatio} estimates optical flow in both spatial and temporal dimensions, and EDVR \cite{wang2019edvr} utilizes deformable convolution to align temporal features.
However, these approaches cannot leverage long-distance temporal features.
To that end, recurrent structures based methods \cite{tao2017detail, haris2019recurrent, isobe2020video, yi2021omniscient, chan2021basicvsr} are proposed for long-distance temporal modeling.
This is often achieved by leveraging the hidden states to build connections across video frames.
In particular, BasicVSR \cite{chan2021basicvsr} introduces a bidirectional recurrent structure that fuses forward and backward propagation features, demonstrating impressive progress.
BasicVSR++ \cite{chan2022basicvsr++} advances BasicVSR \cite{chan2021basicvsr} by introducing the optical flow to the bidirectional recurrent structure.
% the DCN and optical flow are also combined in BasicVSR++ \cite{chan2022basicvsr++} as the flow-guided deformable alignment.
Recently, with Transformer-based approaches \cite{zeng2021improving,cao2021video} being applied in VSR, great successes are achieved by using different attention modules \cite{fu2017look,zheng2017learning} to capture temporal features.
In this paper, instead of developing new architectures, we propose to advance VSR by designing a novel plug-in module to use the prior knowledge.

\paragraph{Prior in Restoration}
Image priors that capture various statistics of natural images have been widely developed and adopted in the field of image restoration (IR). Specifically, for different IR tasks, priors are designed based on the characteristics of the imaging and degradation models.
For example, in the image super-resolution task, the self-similarity prior has been found effective in producing visually appealing results without extensive training on external data, due to the recurrence of natural images within and across scales \cite{ebrahimi2007solving,glasner2009super,freedman2011image,singh2015super,huang2015single}.
On the other hand, the heavily-tailed gradient prior \cite{shan2008high}, sparse kernel prior \cite{fergus2006removing}, $l_0$ gradient prior \cite{xu2013unnatural}, normalized sparsity prior \cite{krishnan2011blind} and dark channel prior \cite{pan2016blind} have been proposed to solve deblurring tasks.
Furthermore, for image denoising tasks, assumptions on prior distribution such as smoothness, low rank and self-similarity have been exploited, leading to the development of methods such as total variation \cite{rudin1992nonlinear}, wavelet-domain processing \cite{donoho1995noising} and BM3D \cite{dabov2007image}.
To address image dehazing problems, existing algorithms often make assumptions on atmospheric light, transmission maps, or clear images. 
For instance, He \etal \cite{he2010single} propose a dark channel prior based on statistical properties of clear images to estimate the transmission map.
Despite the success of these hand-crafted priors, recent research has sought to capture richer image statistics using deep learning models.
For instance, deep priors, \egno, DIP \cite{ulyanov2018deep}, SinGAN \cite{shaham2019singan}, TNRD \cite{chen2016trainable}, and LCM \cite{athar2018latent}, have demonstrated their effectiveness in IR tasks, but their applicability may be limited to the powerless pre-trained models, which are trained on inadequate data.
Recently, the first large-scale foundation model for segmentation -- segment anything model (SAM) is proposed, with 636 million parameters trained on billion-level masks, enabling remarkable representation capability.
This encourages us to leverage the powerful SAM-based prior for VSR enhancement. 

\paragraph{Large-scale Foundation Models}
Large-scale foundation models \cite{devlin2018bert,joshi2020spanbert,radford2019language,kirillov2023segment} have gained great success in both natural language processing (NLP) and computer vision (CV).
Generally, a foundation model is trained with pre-defined tasks, empowering it to transfer in a zero-shot way or accommodate downstream tasks with efficient fine-tuning.
Specifically, BERT \cite{devlin2018bert} is pre-trained with mask-and-predict tasks on a large amount of language data, while GPT-3 \cite{brown2020language} adopts multi-task pre-training with super-large datasets and the model size of 175 billion parameters.
In contrast, CLIP \cite{radford2021learning} contrastively pre-trains on language and vision modalities, enabling language-prompted image classification.
Taking the inspiration of the promptable training fashion of CLIP, SAM \cite{kirillov2023segment} pre-trains a segmentation model that achieves the interactive segmentation given prompts.
Learning on 1 billion masks on 11 million images with 636 million parameters (ViT-H \cite{dosovitskiy2020image}) enables SAM to segment all possible objects existing in one image without being affected by the quality of the image.
We make use of this property to design a novel SAM-guided refinement module to advance existing VSR methods effectively.

\paragraph{Efficient Tuning}
Large-scale foundation models \cite{radford2021learning,kirillov2023segment} learned on abundant training data, \egno, WebImageText \cite{radford2021learning} and SA-1B \cite{kirillov2023segment}, have demonstrated remarkable performance on zero-shot transfer.
However, the training of such foundation models is computationally intensive and requires a significant amount of resources.
One common way to advance the performance of these models on downstream tasks is efficient tuning by introducing a few new parameters to the original model.
That is, only the newly added parameters, \egno, in prompt \cite{jiang2020can,zhou2022learning,zhou2022conditional,jia2022visual} / adapter-based \cite{gao2021clip,zhang2022tip} / residual \cite{yu2022task} tuning manner, are updated during the efficient tuning.
Specifically, prompt tuning often adds new trainable tokens in the input space \cite{jiang2020can,zhou2022learning,zhou2022conditional} or Transformer layers \cite{jia2022visual} for model adaptation, but it is restricted to textual inputs or Transformer architectures.
In contrast, adapter-based \cite{gao2021clip,zhang2022tip} tuning designs new modules added in the middle or the output of a pre-trained network, enabling more flexibility.
Recent proposed residual tuning \cite{yu2022task} opens up a new avenue for efficient tuning by simply adding new tunable parameters to the original ones.
% In this paper, we demonstrate that both fine-tuning and efficient tuning on the proposed module can enhance the performance of VSR.
% This experiment gives the insight to trade off the performance and training parameters.
In this paper, we have shown that incorporating our proposed SEEM module through both fine-tuning and efficient tuning can significantly enhance the performance of VSR. 
Our experiment sheds light on the trade-off between performance and the number of training parameters.

\section{Methodology}

\subsection{Overview}
Inspired by the impressive capability of segment anything model (SAM) to generate high-quality semantic masks without being affected by the image degradation, we propose our method, which leverages this semantic information for VSR.
In this section, we introduce how to get SAM-based representation first, followed by detailing the architecture of our proposed SAM-guided refinement module (SEEM) and the way of applying SEEM to two types of main-stream VSR methods: sliding window-based \cite{kim20183dsrnet,tian2020tdan,wang2019edvr}, and bidirectional recurrent structure-based \cite{tao2017detail, haris2019recurrent, isobe2020video, yi2021omniscient, chan2021basicvsr}.
% and some valuable discussions.

\subsection{SAM-based Representation}
Segment anything model (SAM) \cite{kirillov2023segment} is a large-scale foundation model for image segmentation.
It differs from general segmentation models in two aspects: (i) it is designed and trained in a promptable fashion, enabling the object segmentation given a simple prompt, \egno, a point / a bounding box / a region mask on an image; (ii) the training is implemented on the current largest segmentation dataset (also built by \cite{kirillov2023segment}) with over 1 billion masks on 11 million images in an annotation-and-training loop.

Specifically, SAM consists of an image encoder, a prompt encoder and a mask decoder.
Given an image and a corresponding prompt, they are first encoded by two types of encoders and then decoded by the mask decoder.
We notice that SAM is designed to segment interactively, \ieno, the object segmentation is instructed by a user prompt.
Nevertheless, one can obtain all possible object masks by giving a set of promptable points that cover adequate regions.
Note that given multiple points belonging to the same object, SAM will produce one object mask.
In the paper, we acquire the masks for all objects existing in the image given a $N_g \times N_g$ grid with one point in each grid.
% NxN: 8x8
The shape of the resulting masks is $\mathbb{R}^{C \times H \times W}$, where $C$ is the number of masks.

With the SAM-based prior $M_{sam} \in \mathbb{R}^{C \times H \times W}$, we first concatenate it with its corresponding frame, followed by passing through a feature combination network to get the SAM-based representation $R_{sam} \in \mathbb{R}^{D \times H \times W}$.
We formulate the process as follows,
\begin{equation}\label{eq:sam_r}
    R_{sam} = f_r(concat(I_t, M_{sam})),
\end{equation}
where $f_r$ represents the feature combination network consisting of two convolutional layers and one residual block.
The illustration of this process is given in Figure \ref{fig:framework} (a).

\begin{figure}[ht]
    \centering
    \includegraphics[width=0.4\textwidth]{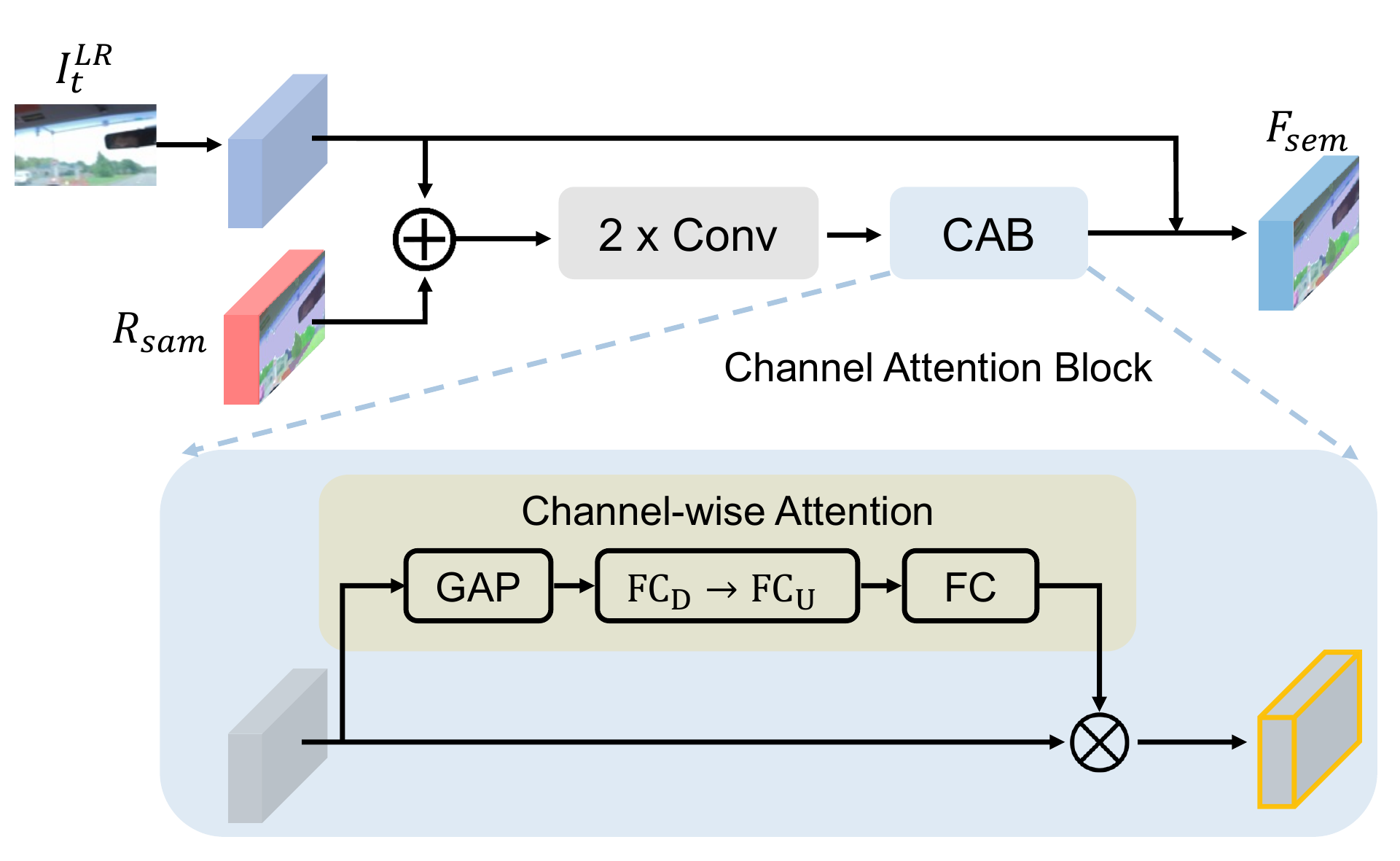}
    %\vspace{-4mm}
    \caption{The structure of SAM-guided refinement module (SEEM). $R_{sam}$: SAM-based representation. $F_{sem}$: semantic-aware feature. GAP: global average pooling. FC: fully-connected layer. $FC_D$: downsampling FC. $FC_U$: upsampling FC. $\otimes$: Hadamard product. $\oplus$: concatenation.}
    \label{fig:seem}
\end{figure}

\subsection{SAM-guided Refinement Module}
SAM-guided refinement module (SEEM) is designed to combine the SAM-based representation $R_{sam}$ and the feature of the corresponding frame to produce a semantic-aware feature for enhanced alignment and fusion, which is shown in Figure \ref{fig:seem}.
Specifically, given the feature of a frame and its corresponding SAM-based representation, they are first concatenated as the input for two convolutional layers.
The mapped feature will forward through a channel attention block (CAB) \cite{zhang2018image} to explore the interdependencies among channels for adaptively rescaling channel-wise features.
Concretely, with the mapped feature $F_m \in \mathbb{R}^{D \times H \times W}$, the channel-wise attention is computed as follows,
\begin{equation}
    \begin{split}
        F_p & = GAP(F_m), F_p \in \mathbb{R}^{D \times 1 \times 1}, \\
        F_d & = \phi_D(F_p), F_d \in \mathbb{R}^{\frac{D}{r} \times 1 \times 1}, \\
        F_u & = \phi_U(F_d), F_u \in \mathbb{R}^{D \times 1 \times 1}, \\
        F_a & = \phi(F_u), F_a \in \mathbb{R}^{D \times 1 \times 1},
    \end{split}
\end{equation}
where $GAP$ is the global average pooling and $\phi$ is the fully-connected layer.

The channel-wise attention $F_a$ is then multiplied with the feature $F_m$ to get the output of CAB as defined below,
\begin{equation}
    F_{cab} = F_a \odot F_m, F_{cab} \in \mathbb{R}^{D \times H \times W},
\end{equation}
where $\odot$ is the Hadamard product.

The output of our SEEM is the semantic-aware feature, which is obtained by summing up the $F_{cab}$ and $F_m$, \ieno, $F_{sem} = F_m + F_{cab}$.

\subsection{Apply SEEM to Sliding Window-based Methods}
\subsubsection{Revisiting EDVR \cite{wang2019edvr}}
EDVR \cite{wang2019edvr} is a typical method based on sliding window, which takes 2N+1 low-resolution frames as inputs and generates one high-resolution output.
EDVR aligns 2N neighboring frames with the reference frame located in the middle using an alignment module. 
The aligned features are then combined through a fusion module.
Finally, the reconstruction module digests the fused features and outputs one high-resolution frame.

To be specific, EDVR has proposed two key modules: the PCD alignment module and the TSA fusion module.
PCD alignment module advances the normal deformable convolution (DCN) \cite{tian2020tdan} with pyramidal processing \cite{ranjan2017optical,sun2018pwc} and cascading refinement \cite{hui2018liteflownet,hui2020lightweight}.
For simplicity, we give a brief introduction of how DCN works instead of the detailed pyramid and cascading.
Concretely, DCN is comprised by two modules: deformable alignment module (DAM) and deformable convolution module (DCM).
Given the features of the target frame and the reference frame $F_{t+i}, F_t (i \in[-N:+N])$ as the inputs, DAM aims to generate sampling parameters $\Theta$ for $F_i$ by the following equation,
\begin{equation}\label{eq:dam}
    \Theta = f_{dam}(F_{t+i}, F_t), \Theta=\{\Delta p_n\},
\end{equation}
where $\Delta p_n$ is the offset of the convolutional kernel and $n$ is the number of kernels.
Taking the $\Theta$ and $F_t$ as the inputs for DCM, we can obtain the aligned feature $F_{t+i}^{\star}$, as defined below,
\begin{equation}
    F_{t+i}^{\star} = f_{dcm}(F_{t+i}, \Theta).
\end{equation}

The aligned features $\{F_{t+i}^{\star} | i \in[-N:+N] \}$ are passing through the TSA fusion module, which considers both temporal and spatial attention.
The fusion process is formulated as follows,
\begin{equation}\label{eq:tsa}
    F_{fuse} = f_{tsa}(F_{t+i}^{\star} | i \in[-N:+N]).
\end{equation}

Finally, the fused feature $F_{fuse}$ is used to reconstruct the high-resolution frame, as shown below,
\begin{equation}
    I_t^{hr} = f_{rec}(F_{fuse}),
\end{equation}
where $I_t^{hr}$ is the recovered high-resolution frame and $f_{rec}$ is the reconstruction module.

\subsubsection{Equip EDVR with SEEM}
We apply our SEEM to boost the performance of EDVR in alignment, fusion and reconstruction, which is shown in Figure \ref{fig:framework} (b).
Specifically, for alignment, we make the feature of reference frame be semantic-aware such that the Eq. \ref{eq:dam} in EDVR is re-formulated as follows,
\begin{equation}
    \Theta = f_{dam}(F_{t+i}, f_{seem}(F_t, R_{sam}) + F_t),
\end{equation}
where $R_{sam}$ is computed in Eq. \ref{eq:sam_r} and $f_{seem}$ is our SEEM.

We further incorporate our SEEM to facilitate both the fusion and reconstruction processes.
The formulation is shown below,
\begin{equation}
    \begin{split}
        F_{rec} & = f_{rec}(f_{seem}(F_{fuse}, R_{sam}) + F_{fuse}), \\
        F_{rec}^{'} & = f_{seem}(F_{rec}, R_{sam}) + F_{rec}.
    \end{split}
\end{equation}

\paragraph{The Design of SEEM}
Our SEEM is specifically designed in two aspects: (i) it is an easy plug-in module that can be seamlessly integrated into the existing methods without altering the original architecture, enabling its generalization and scalability; and (ii) the output of the SEEM is added to the current feature as a residual, a technique that has been shown to be empirically effective \cite{he2016deep,yu2022task,gao2021clip}, which can preserve the old knowledge from the pre-trained model as well as explore the task-specific knowledge.
Furthermore, SEEM can be trained in a parameter-efficient tuning manner, \ieno, only the parameters of SEEM are updated during fine-tuning.

\subsection{Apply SEEM to Bidirectional Recurrent Structure-based Methods}
\subsubsection{Revisiting BasicVSR \cite{chan2021basicvsr}}
We focus on refining a representative method -- BasicVSR, among bidirectional recurrent structure-based methods, as an example to manifest how SEEM works in this kind of pipeline.
First, we provide a brief overview of how BasicVSR functions.
BasicVSR has proposed two key components: bidirectional propagation and flow-based feature-level alignment.
Specifically, given a low-resolution frame $I_t$ and its neighboring frames $I_{t-1}$ and $I_{t+1}$, the outputs of forward ($f_F$) and backward ($f_B$) branches are:
\begin{equation}
    \begin{split}
        h_t^b & = f_B(I_t, I_{t+1}, h_{t+1}^b), \\
        h_t^f & = f_F(I_t, I_{t-1}, h_{t-1}^f).
    \end{split}
\end{equation}

The output features are then used in the feature-level alignment. 
To align the features, the optical flow is first estimated and leveraged for spatial warping.
This process can be formulated as follows,
\begin{equation}
    \begin{split}
        s_t^b & = Flow(I_t, I_{t+1}), \\
        s_t^f & = Flow(I_t, I_{t-1}), \\
        \hat{h}_t^b & = Wrap(h_{t \pm 1}^b, s_t^b), \\
        \hat{h}_t^f & = Wrap(h_{t \pm 1}^f, s_t^f), \\
    \end{split}
\end{equation}
where $Flow$ is the optical flow estimate model and $Wrap$ is the warping operation.
After the feature warping, they also adopt several residual blocks for refinement, as shown below,
\begin{equation} \label{eq:refine}
    \begin{split}
        \tilde{h}_t^b & = Ref_b(\psi (I_t, \hat{h}_t^b)), \\
        \tilde{h}_t^f & = Ref_f(\psi (I_t, \hat{h}_t^f)), \\
    \end{split}
\end{equation}
where $Ref_{\{b,f\}}$ represents a stack of residual blocks and $\psi$ is the convolutional layer. The $\tilde{h}_t^b$ and $\tilde{h}_t^f$ are then forwarded to the reconstruction module for the generation of high-resolution frames.

\subsubsection{Equip BasicVSR with SEEM}
Our SEEM is integrated into BasicVSR with two purposes: refinement of the warping feature and enhancement of the representation before reconstruction.
To be specific, given the current frame $I_t$ and the wrapped features $\hat{h}_t^b, \hat{h}_t^f$, Eq. \ref{eq:refine} is re-formulated as follows with the refinement,
\begin{equation}
    \begin{split}
        \tilde{h}_t^b & = Ref_b(f_{seem}(\psi (I_t, \hat{h}_t^b), R_{sam}) + \psi (I_t, \hat{h}_t^b)), \\
        \tilde{h}_t^f & = Ref_b(f_{seem}(\psi (I_t, \hat{h}_t^f), R_{sam}) + \psi (I_t, \hat{h}_t^f)).
    \end{split}
\end{equation}

The refined feature is enhanced by SEEM again, 
\begin{equation}
    \begin{split}
        \tilde{h}_t^{b'} & = f_{seem}(\tilde{h}_t^b, R_{sam}) + \tilde{h}_t^b, \\
        \tilde{h}_t^{f'} & = f_{seem}(\tilde{h}_t^f, R_{sam}) + \tilde{h}_t^f. \\
    \end{split}
\end{equation}

\section{Experiments}

\subsection{Datasets and Evaluation}

\paragraph{REDS \cite{nah2019ntire}} It is an NTIRE19 challenge dataset including 300 video sequences, which have been split into three sets for training, validation, and testing purposes. 
Specifically, there are 240 sequences designated for training, 30 sequences for validation, and an additional 30 sequences for testing.
Each video sequence in the dataset consists of 100 frames, each with a resolution of 1280 $\times$ 720.
The protocol of splitting training and test sets is following past works \cite{li2020mucan,wang2019edvr,chan2021basicvsr}, \ieno, selecting four sequences\footnote{Clips 000,011,015,020 of the REDS training set.} as the testing set which is called {\it REDS4}.
For the training set, it is formed by the remaining 266 sequences from the combined training and validation sets.
\begin{table*}[ht]
  \caption{The quantitative comparison on REDS4 \cite{nah2019ntire} dataset for $4\times$ VSR. The results are evaluated on RGB channels.}
  \label{tab:reds}
  % \vspace{-4mm}
  \adjustbox{width=0.9\textwidth}{
  \centering
  \begin{tabular}{ l | l| c c| c c| c c| c c| c c}
    \toprule
    \multirow{2}{*}{Category} & \multirow{2}{*}{Method} &  \multicolumn{2}{c}{Clip\_000} &  \multicolumn{2}{c}{Clip\_011} &  \multicolumn{2}{c}{Clip\_015} &  \multicolumn{2}{c}{Clip\_020}  &  \multicolumn{2}{c}{Average}  \\
    & & PSNR $\uparrow$ & SSIM $\uparrow$ & PSNR $\uparrow$ & SSIM $\uparrow$ & PSNR $\uparrow$ & SSIM $\uparrow$ & PSNR $\uparrow$ & SSIM $\uparrow$ & PSNR $\uparrow$ & SSIM $\uparrow$ \\
    \midrule
    \multirow{3}{*}{SISR-based} & Bicubic & 24.5500 & 0.64890 & 26.0600 & 0.72610 & 28.5200 & 0.80340 & 25.4100 & 0.73860 & 26.1400 & 0.72920 \\
    & RCAN~\cite{zhang2018image} & 26.1700 & 0.73710 & 29.3400 & 0.82550 & 31.8500 & 0.88810 & 27.7400 & 0.82930 & 28.7800 & 0.82000  \\
    & CSNLN~\cite{mei2020image} & 26.1700 & 0.73790 & 29.4600 & 0.82600 & 32.0000 & 0.88900 & 27.6900 & 0.82530 & 28.8300 & 0.81960  \\
    \midrule
    \multirow{5}{*}{SW-based} & TOFlow~\cite{xue2019video} & 26.5200 & 0.75400 & 27.8000 & 0.78580 & 30.6700 & 0.86090 & 26.9200 & 0.79530 & 27.9800 & 0.79900 \\
    & DUF~\cite{jo2018deep} & 27.3000 & 0.79370 & 28.3800 & 0.80560 & 31.5500 & 0.88460 & 27.3000 & 0.81640 & 28.6300 & 0.82510  \\
    & EDVR \cite{wang2019edvr} & 27.7464 & 0.81529 & 31.2897 & 0.87323 & 33.4808 & 0.91326 & 29.5926 & 0.87762 & 30.5274 & 0.86985 \\
    & \CC  & \CC 27.7738 & \CC 0.81713 & \CC 31.3322 & \CC 0.87412 & \CC 33.4959 & \CC 0.91374 & \CC 29.6091 & \CC 0.87817 & \CC 30.5528 & \CC 0.87079 \\
    & \multirow{-2}{*}{\CC EDVR + Ours} & \CC {\bf +0.0274} & \CC {\bf +0.00184} & \CC {\bf +0.0425} & \CC {\bf +0.00089} & \CC {\bf +0.0151} & \CC {\bf +0.00048} & \CC {\bf +0.0165} & \CC {\bf +0.00055} & \CC {\bf +0.0254} & \CC {\bf +0.00094} \\
    \midrule
    \multirow{2}{*}{BRS-based} & BasicVSR \cite{chan2021basicvsr} & 28.4004 & 0.84342 & 32.4675 & 0.89792 & 34.1752 & 0.92239 & 30.6253 & 0.90004 & 31.4171 & 0.89094 \\
    & \CC & \CC 28.4475 & \CC 0.84485 & \CC 32.5952 & \CC 0.89929 & \CC 34.2913 & \CC 0.92402 & \CC 30.6853 & \CC 0.90085 & \CC 31.5048 & \CC 0.89225 \\
    & \multirow{-2}{*}{\CC BasicVSR + Ours} & \CC {\bf +0.0471} & \CC {\bf +0.00143} & \CC {\bf +0.1277} & \CC {\bf +0.00137} & \CC {\bf +0.1161} & \CC {\bf +0.00163} & \CC {\bf +0.0600} & \CC {\bf +0.00081} & \CC {\bf +0.0877} & \CC {\bf +0.00131} \\
    \bottomrule   
  \end{tabular}}
\end{table*}
\begin{figure*}[ht]
    \centering
    \includegraphics[width=0.9\textwidth]{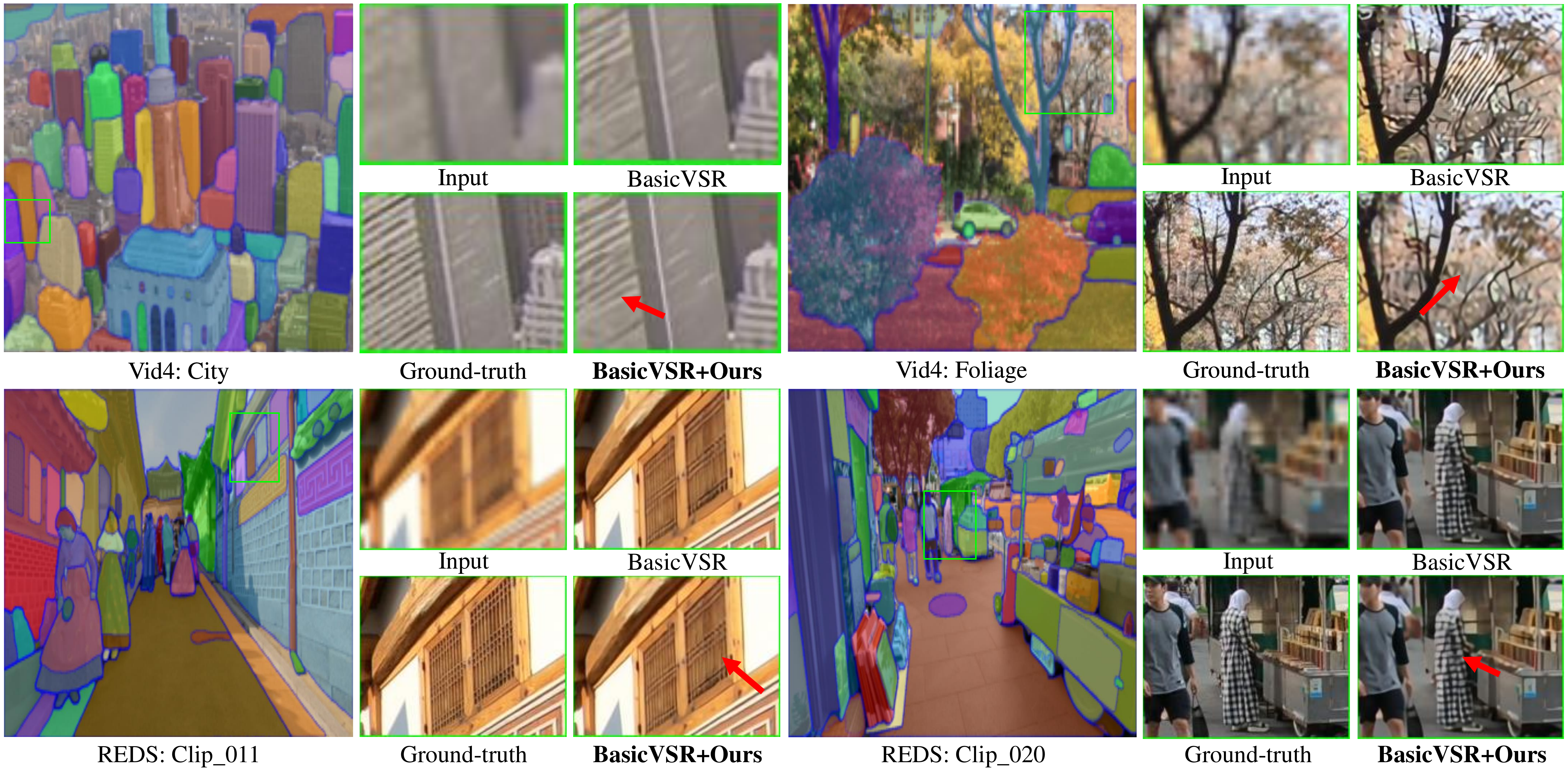}
    %\vspace{-4mm}
    \caption{The qualitative comparison of BasicVSR with and without our SEEM. The top row is tested on Vid4 under two scenes: city and foliage, while the bottom row is evaluated on two clips of REDS. The masks are extracted from the segment anything model. We enlarge the small patches in the images for better observation. Zoom in for more details.}
    \label{fig:basic_vis}
\end{figure*}
\begin{table*}[ht]
  \caption{The performance of tuning our proposed SEEM only on REDS4 \cite{nah2019ntire} dataset for $4\times$ VSR. The results are evaluated on RGB channels. $*$: freezing the parameters of the model.}
  % \vspace{-4mm}
  \label{tab:tuning_reds}
  \adjustbox{width=0.9\textwidth}{
  \centering
  \begin{tabular}{l| c c| c c| c c| c c| c c}
    \toprule
    \multirow{2}{*}{Method} &  \multicolumn{2}{c}{Clip\_000} &  \multicolumn{2}{c}{Clip\_011} &  \multicolumn{2}{c}{Clip\_015} &  \multicolumn{2}{c}{Clip\_020}  &  \multicolumn{2}{c}{Average}  \\
    & PSNR $\uparrow$ & SSIM $\uparrow$ & PSNR $\uparrow$ & SSIM $\uparrow$ & PSNR $\uparrow$ & SSIM $\uparrow$ & PSNR $\uparrow$ & SSIM $\uparrow$ & PSNR $\uparrow$ & SSIM $\uparrow$ \\
    \midrule
    EDVR \cite{wang2019edvr} & 27.7464 & 0.81529 & 31.2897 & 0.87323 & 33.4808 & 0.91326 & 29.5926 & 0.87762 & 30.5274 & 0.86985 \\
    \rowcolor{gray!20}
    \rowcolor{gray!20}
    & 27.7643 & 0.81676 & 31.3171 & 0.87363 & 33.4476 & 0.91322 & 29.5978 & 0.87786 & 30.5317 & 0.87037 \\
    \rowcolor{gray!20}
    \multirow{-2}{*}{EDVR$^{*}$ + Ours} & {\bf +0.0179} & {\bf +0.00147} & {\bf +0.0274} & {\bf +0.00040} & -0.0332 & -0.00004 & {\bf +0.0052} & {\bf +0.00024} & {\bf +0.0043} & {\bf +0.00052}  \\
    \bottomrule   
  \end{tabular}}
\end{table*}

\paragraph{Vimeo-90K \cite{xue2019video}} This is the current largest video super-resolution dataset with 64,612 sequences for training and 7,824 for testing.
Each sequence is comprised of seven individual frames, each with a resolution of 448 $\times$ 256.

\paragraph{Evaluation}
To ensure a fair comparison, we adopt the evaluation protocol used in previous works \cite{chan2021basicvsr} to evaluate the performance of the proposed method with 4 $\times$ downsampling, \ieno, MATLAB bicubic downsample (BI).
As per \cite{chan2021basicvsr,li2020mucan}, the evaluation metrics are: (i) peak signal-to-noise ratio (PSNR) and (ii) structural similarity index (SSIM) \cite{wang2004image}.

\subsection{Implementation Details} 
We choose EDVR \cite{wang2019edvr} and BasicVSR \cite{chan2021basicvsr} as our baselines.
To ensure a fair comparison, we maintain the same network structures as those in the original papers.
The training batch sizes are set to 4 for EDVR and 6 for BasicVSR.
We use an 8$\times$8 grid with points for SAM to generate masks.
Hence, each low-resolution image has a maximum of 64 masks. For those with less than 64 masks, the remaining ones are padded with zeros.
As per EDVR \cite{wang2019edvr} and BasicVSR \cite{chan2021basicvsr}, we use RGB patches of size 64$\times$64 as input.
The augmentations used in training are random horizontal flips and rotations.
Please refer to EDVR \cite{wang2019edvr} and BasicVSR \cite{chan2021basicvsr} for more training details.
We also provide them in supplementary material for convenience.

\begin{table*}[ht]
  \caption{The quantitative comparison on Vimeo90K \cite{xue2019video} testing dataset for $4\times$ VSR. The results are evaluated on RGB channels.}
  % \vspace{-4mm}
  \label{tab:vimeo}
  \adjustbox{width=0.80\textwidth}{
  \centering
  \begin{tabular}{ l| c c| c c| c c| c c}
    \toprule
       \multirow{2}{*}{Method} & \multicolumn{2}{c}{Fast} & \multicolumn{2}{c}{Medium} & \multicolumn{2}{c}{Slow}& \multicolumn{2}{c}{Average} \\
       & PSNR $\uparrow$ & SSIM $\uparrow$ & PSNR $\uparrow$ & SSIM $\uparrow$ & PSNR $\uparrow$ & SSIM $\uparrow$ & PSNR $\uparrow$ & SSIM $\uparrow$ \\
        \midrule
        EDVR \cite{wang2019edvr} & 38.7020 & 0.95437 & 35.9896 & 0.94125 & 32.9052 & 0.91235 & 35.7912 & 0.93739 \\
        \rowcolor{gray!20}
        & 38.7200 & 0.95442 & 36.0384 & 0.94165 & 32.9447 & 0.91282 & 35.8333 & 0.93775 \\
        \rowcolor{gray!20}
        \multirow{-2}{*}{EDVR + Ours} & {\bf +0.0180} & {\bf +0.00005} & {\bf +0.0488} & {\bf +0.00040} & {\bf +0.0395} & {\bf +0.00047} & {\bf +0.0421} & {\bf +0.00036} \\
        BasicVSR \cite{chan2021basicvsr} & 38.2954 & 0.95152 & 35.5435 & 0.93681 & 32.5003 & 0.90629 & 35.3601 & 0.93287  \\
        \rowcolor{gray!20}
         &  38.4074 &  0.95222 &  35.6619 &  0.93777 &  32.6234 &  0.90775 &  35.4786 &  0.93390 \\
        \rowcolor{gray!20}
        \multirow{-2}{*}{BasicVSR + Ours} &  {\bf +0.1120} &  {\bf +0.00070} &  {\bf +0.1185} &  {\bf +0.00096} &  {\bf +0.1231} &  {\bf +0.00146} &  {\bf +0.1184} &  {\bf +0.00102} \\
    \bottomrule
\end{tabular}}
\end{table*}
\begin{figure*}[ht]
    \centering
    \includegraphics[width=0.95\textwidth]{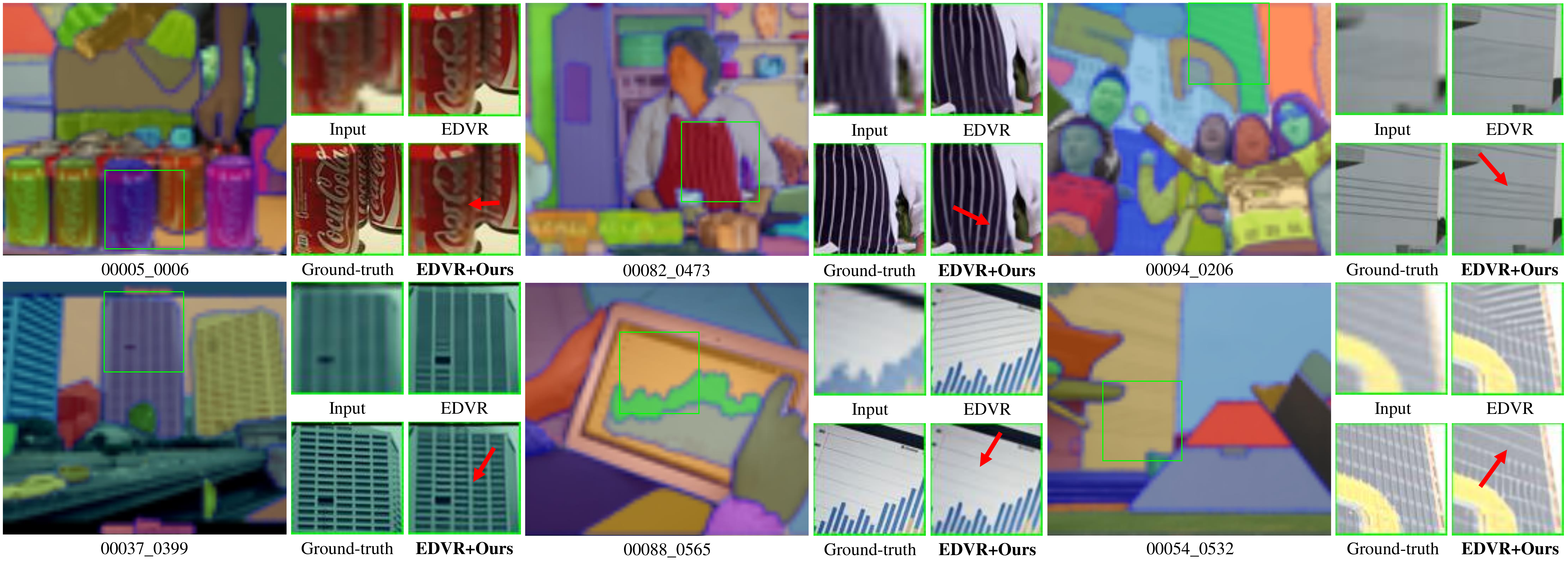}
    %\vspace{-4mm}
    \caption{The qualitative comparison of EDVR with and without our SEEM on Vimeo90K dataset.
    We show six examples for comparing the differences.
    The masks are extracted from the segment anything model. We enlarge the small patches in the images for better observation. Zoom in for more details.}
    \label{fig:edvr_vis}
    % %\vspace{-4mm}
\end{figure*}

\subsection{Compared Methods}
We compare our method with many existing state of the arts.
They can be simply grouped into three categories, namely, single image super-resolution-based (SISR-based) \cite{mei2020image,zhang2018image}, sliding window-based (SW-based) \cite{jo2018deep,wang2019edvr,xue2019video}, and bidirectional recurrent structure-based (BRS-based) \cite{chan2021basicvsr,sajjadi2018frame}.
It is worth noting that we cited the results from the original papers or reproduced the results using the official code released by the authors to ensure a fair comparison.

\begin{table*}[ht]
  \caption{The quantitative comparison on Vid4 testing dataset \cite{caballero2017real} for $4\times$ VSR. The results are evaluated on Y-channel.}
  % \vspace{-4mm}
  \label{tab:vid4}
  \adjustbox{width=0.9\textwidth}{
  \centering
  \begin{tabular}{ l| c c| c c| c c| c c| c c}
    \toprule
       \multirow{2}{*}{Method} & \multicolumn{2}{c}{Calendar} & \multicolumn{2}{c}{City} & \multicolumn{2}{c}{Foliage} & \multicolumn{2}{c}{Walk} & \multicolumn{2}{c}{Average} \\
       & PSNR $\uparrow$ & SSIM $\uparrow$ & PSNR $\uparrow$ & SSIM $\uparrow$ & PSNR $\uparrow$ & SSIM $\uparrow$ & PSNR $\uparrow$ & SSIM $\uparrow$ & PSNR $\uparrow$ & SSIM $\uparrow$ \\
        \midrule
        DRVSR \cite{tao2017detail} & 22.8800 & 0.75860 & 27.0600 & 0.76980 & 25.5800 & 0.73070 & 29.1100 & 0.88760 & 25.5200 & 0.76000 \\
        FRVSR \cite{sajjadi2018frame} & 23.4600 & 0.78540 & 27.7000 & 0.80990 & 25.9600 & 0.75600 & 29.6900 & 0.89900 & 26.6900 & 0.82200 \\
        MMCNN \cite{wang2018multi} & 23.6300 & 0.79690 & 27.4700 & 0.80830 & 26.0100 & 0.75320 & 29.9400 & 0.90300 & 26.2800 & 0.78440 \\
        MTUDM \cite{yi2019multi} & 23.7600 & 0.80260 & 27.6700 & 0.81450 & 26.0800 & 0.75870 & 30.1600 & 0.90690 & 26.5700 & 0.79890 \\
        DUF \cite{jo2018deep} & 23.8500 & 0.80520 & 27.9700 & 0.82530 & 26.2200 & 0.76460 & 30.4700 & 0.91180 & 27.3400 & 0.83270 \\
        % RBPN \cite{haris2019recurrent}         & 24.33 & 0.8244 & 28.28 & 0.8413 & 26.46 & 0.7753 & 30.58 & 0.9130 & 27.16 & 0.8190 \\
        % PFNL \cite{yi2019progressive}         & 24.37 & 0.8246 & 28.09 & 0.8385 & 26.51 & 0.7768 & 30.64 & 0.9134 & 27.40 & 0.8384 \\
        % FFCVSR \cite{yan2019frame}     & 24.39 & 0.8250 & 27.80 & 0.8314 & 26.70 & 0.7868 & 30.55 & 0.9124 & 26.97 & 0.8300 \\
        % RLSP7-256 \cite{fuoli2019efficient}    & 24.60 & 0.8335 & 28.14 & 0.8453 & 30.88 & 0.9192 & 27.60 & 0.8476 & 27.55 & 0.8476 \\
        TDAN \cite{tian2020tdan} & 23.5600 & 0.78960 & 27.5300 & 0.80280 & 26.0000 & 0.74910 & 29.9900 & 0.90320 & 26.8600 & 0.81400 \\
        \midrule
        EDVR \cite{wang2019edvr} & 24.0483 &  0.81475 & 27.9973 & 0.81219 & 26.3358 & 0.76352 & 31.0227 &  0.91516 & 27.3510 &  0.82640\\
        \rowcolor{gray!20}
        & 24.0544 & 0.81555 & 28.0030 & 0.81261 & 26.3398 & 0.76396 & 31.0458 & 0.91554 & 27.3608 & 0.82692 \\
        \rowcolor{gray!20}
        \multirow{-2}{*}{EDVR + Ours} & {\bf +0.0061} & {\bf +0.00080} & {\bf +0.0057} & {\bf +0.00042} & {\bf +0.0040} & {\bf +0.00044} & {\bf +0.0231} & {\bf +0.00038} & {\bf +0.0098} & {\bf +0.00052} \\
        BasicVSR \cite{chan2021basicvsr} & 23.8681 & 0.80940 & 27.6571 & 0.80501 & 26.4727 & 0.77104 & 30.9555 & 0.91476 & 27.2383 & 0.82505 \\
        \rowcolor{gray!20}
        & 23.9438 & 0.81162 & 27.6904 & 0.80633 & 26.5838 & 0.77531 & 31.0863 & 0.91659 & 27.3260 & 0.82746 \\
        \rowcolor{gray!20}
        \multirow{-2}{*}{BasicVSR + Ours} & {\bf +0.0757} & {\bf +0.0022} & {\bf +0.0332} & {\bf +0.0013} & {\bf +0.1110} & {\bf +0.0043} & {\bf +0.1308} & {\bf +0.0018} & {\bf +0.0877} & {\bf +0.0024} \\
    \bottomrule
\end{tabular}}
\end{table*}

\subsection{Main Results}
\subsubsection{Results on REDS4 \cite{nah2019ntire}}
In Table \ref{tab:reds}, we compare our method with two typical baselines as well as other existing methods.
Overall, our SEEM can boost the baseline methods in all scenarios, \ieno, four testing clips in REDS.
% Moreover, some additional observations can be made.
Regarding the averaged performance on four clips, our SEEM improves EDVR by {\bf 0.0254} dB / {\bf 0.00094} on PSNR / SSIM and BasicVSR by {\bf 0.0877} dB / {\bf 0.00131} on PSNR / SSIM.
The consistent performance gains on various methods manifest that the utilization of the robust semantic-aware prior knowledge via our SEEM indeed enhances the quality of VSR.
We also present qualitative comparisons in Figure \ref{fig:basic_vis} (bottom row), which demonstrate that our SEEM can significantly enhance the visual quality of the results compared with baseline methods.
 For example, our SEEM can improve the sharpness of the window grille in the bottom-left image, as well as the texture of the plaid pattern on the skirt in the bottom-right image.

Taking the inspiration of efficient tuning on pre-trained large-scale foundation models, we investigate the ability of our SEEM in a similar way.
That is, with the parameters of a pre-trained VSR model being frozen, we train our SEEM only for performance improvement.
The experimental results are shown in Table \ref{tab:tuning_reds}.
We found that this efficient tuning can boost both PSNR and SSIM even though fewer parameters are updated.
This offers a practical alternative that can overcome the constraints of limited storage capacity.

\subsubsection{Results on Vimeo-90K \cite{xue2019video}}
We also conduct the evaluation on Vimeo-90K dataset \cite{xue2019video}, which is shown in Table \ref{tab:vimeo}.
% as the training set while Vimeo-90K-T \cite{xue2019video} as the test set.
Again, we found that our SEEM can always improve the baseline methods, often by large margins.
Concretely, the performance gains from adding our SEEM on EDVR and BasicVSR are {\bf 0.0421} dB / {\bf 0.00036} (PSNR / SSIM) and {\bf 0.1184} dB / {\bf 0.00102} (PSNR / SSIM).
Moreover, we have observed that our SEEM demonstrates superiority on the most challenging slow-motion videos though facing a slight performance drop on fast-motion videos.

\subsubsection{Results on Vid4 \cite{liu2013bayesian}}
To further verify the generalization capabilities of the proposed method, we train the model using the Vimeo-90K dataset \cite{xue2019video} and evaluate its performance on the Vid4 dataset \cite{liu2013bayesian}, as shown in Table \ref{tab:vid4}.
% providing a rigorous assessment of its ability to generalize to other datasets beyond its training data.
We found that all baseline methods benefit from adding our SEEM, which indicates that incorporation with semantic information not only improves the performance of a VSR method within a domain (see REDS results in Table \ref{tab:reds}) but also advances the generalization ability of the applied method across domains.
Apart from the quantitative results, we also show some visual examples in Figure \ref{fig:basic_vis} (top row).
We found that in the foliage image of the Vid4 dataset (up-right image), BasicVSR yields obvious artifacts while with our SEEM the artifacts disappear.

\subsection{Further Analysis}\label{subsec:ablation}

\paragraph{The Effect of Using SEEM}
As our method is fine-tuned on a pre-trained VSR method, we additionally compare the performance of directly fine-tuning the method to fine-tuning with both the method and our SEEM incorporated in Table \ref{tab:no_tune_ave}.
We found that direct fine-tuning EDVR or BasicVSR does not benefit anymore while adding our SEEM for fine-tuning brings significant improvements, \egno, {\bf 0.0148} dB / {\bf 0.00084} (PSNR / SSIM) for EDVR and {\bf 0.1038} dB / {\bf 0.00601} (PSNR / SSIM) for BasicVSR.
The full results regarding various scenes on Vid4 are provided in the supplementary material.

\begin{table}[ht]
  \caption{The comparison of fine-tuning baseline methods with and without our SEEM on Vid4 dataset for $4\times$ VSR. The results are evaluated on Y-channel.}
  % \vspace{-4mm}
  \label{tab:no_tune_ave}
  \adjustbox{width=0.27\textwidth}{
  \centering
  \begin{tabular}{l|cc}
    \toprule
       \multirow{2}{*}{Method} & \multicolumn{2}{c}{Vid4 Average} \\
        & PSNR $\uparrow$ & SSIM $\uparrow$ \\
        \midrule
        EDVR \cite{wang2019edvr} & 27.3460 & 0.82608 \\
        \rowcolor{gray!20}
        & 27.3608 & 0.82692 \\
        \rowcolor{gray!20}
        \multirow{-2}{*}{EDVR + Ours} & {\bf +0.0148} & {\bf +0.00084} \\
        BasicVSR \cite{chan2021basicvsr} & 27.2222 & 0.82145 \\
        \rowcolor{gray!20}
        & 27.3260 & 0.82746 \\
        \rowcolor{gray!20}
        \multirow{-2}{*}{BasicVSR + Ours} & {\bf +0.1038} & {\bf +0.00601} \\
    \bottomrule
\end{tabular}}
\end{table}

\paragraph{Applying SEEM to Different Branches}
As mentioned in the Method section, the proposed SEEM is inserted into BasicVSR at two branches.
Here, we perform an ablation study to analyze the impact of integrating SEEM at different branches, as shown in Table \ref{tab:basicvsr_fb}.
We observed that adding SEEM to either branch can lead to improvements, while adding it to both branches achieves the best performance.
We further provide the full results in the supplementary material.
\begin{table}[ht]
  \caption{The ablation study of adding SEEM to different branches on BasicVSR \cite{chan2021basicvsr} with $4\times$ upsampling. The results are evaluated on Y-channel.}
  % \vspace{-4mm}
  \label{tab:basicvsr_fb}
  \adjustbox{width=0.40\textwidth}{
  \centering
  \begin{tabular}{l|cc|cc}
    \toprule
    \multirow{2}{*}{Method} & \multirow{2}{*}{Forward} & \multirow{2}{*}{Backward} & \multicolumn{2}{c}{Vid4 Average} \\
    & & & PSNR $\uparrow$ & SSIM $\uparrow$ \\
    \midrule
    \multirow{4}{*}{BasicVSR \cite{chan2021basicvsr}} & & & 27.2383 & 0.82505 \\
    & $\checkmark$ & & 27.2560 & 0.82360   \\
    & & $\checkmark$ & 27.2546 & 0.82272 \\
    & $\checkmark$ & $\checkmark$ & {\bf 27.3260} & {\bf 0.82746} \\
    \bottomrule    
  \end{tabular}}
\end{table}

\begin{figure}
    \centering
    \includegraphics[width=0.44\textwidth]{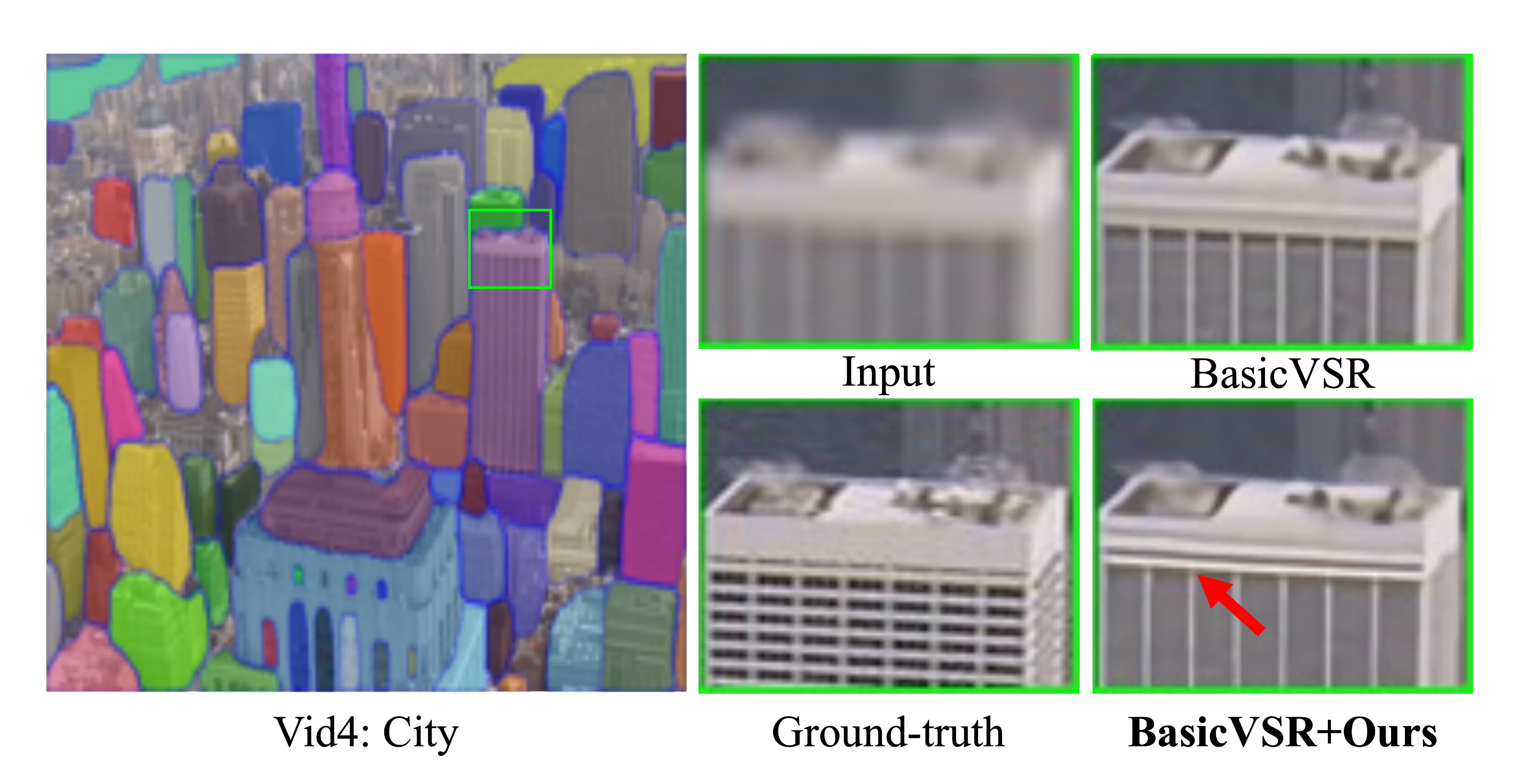}
    %\vspace{-4mm}
    \caption{The failure case with our SEEM being applied on BasicVSR under city scene on Vid4. Zoom in for better observation.}
    \label{fig:failure}
    %\vspace{-4mm}
\end{figure}

\paragraph{Failure Cases}
We further present failure cases to investigate the limitations of our proposed method.
In Figure \ref{fig:failure}, we found that our SEEM tends to introduce unseen artifacts to the generated high-resolution frame, despite the overall quality of the images being visually appealing to humans.

\section{Limitations}
Due to the increasing popularity of large-scale foundation models, it has been a trend to leverage their abilities to advance downstream tasks.
We explore the knowledge of segment anything model (SAM) to improve the quality of video super-resolution (VSR) by proposing a simple plug-in module -- SEEM.
While we utilize the masks generated from SAM in our approach, it is worth noting that this is just one possible approach and there may be other ways to achieve similar results.
For example, one can explore the prior knowledge embedded within the model, \egno, feature-level information.
In addition, our approach is broader than improving VSR.
We will investigate its ability on other computer vision tasks as future work.

\section{Conclusion}
In this paper, we found the limitations of existing video super-resolution (VSR) methods, \ieno, (i) the lack of leveraging valuable semantic information to enhance the quality of VSR; (ii) the used prior knowledge is imprecise when facing image degradation.
To that end, we have explored a more robust and semantic-aware prior, which is obtained from a powerful large-scale foundation model -- segment anything model (SAM).
To utilize the SAM-based prior, we have proposed a simple yet effective module -- SAM-guided refinement module (SEEM), enabling better alignment and fusion by using semantic information.
Specifically, our SEEM is designed as a plug-in that can be seamlessly integrated into existing VSR methods for improved performance.
The capability of SEEM stems from the attention-based combination of extracted features and SAM-based representation.
We verify its generalization and scalability by applying it to two representative VSR methods: EDVR and BasicVSR, on three widely-used datasets (Vimeo-90K, REDS and Vid4), leading to consistent performance gains.
Furthermore, we found that SEEM can be used in a parameter-efficient tuning manner for a performance boost.
% \ieno, only the parameters of SEEM are trainable during fine-tuning.
This provides greater flexibility for adjusting the trade-off between performance and the number of training parameters.
% \clearpage
% \section{CCS Concepts and User-Defined Keywords}
\bibliographystyle{ACM-Reference-Format}
\bibliography{acmart.bib}
 
% \appendix
% \section{appendix}
\end{document}